\documentclass{article}
\usepackage{amssymb}
\usepackage{amsmath}
\usepackage{graphicx}
\usepackage{enumitem}
\usepackage{booktabs}
\usepackage{float}
\usepackage{multirow}
\usepackage[table,xcdraw]{xcolor}
\usepackage{todonotes}
\usepackage{tabularx}
\usepackage{array}
\usepackage{wrapfig}
\usepackage[font=small]{caption}
\newcommand{\equalcontrib}{\textsuperscript{*}}
\newcolumntype{Y}[1]{>{\centering\arraybackslash}p{#1}} % fixed width column

\usepackage[preprint]{corl_2025} % Uncomment for pre-prints (e.g., arxiv); This is like ``final'', but will remove the CORL footnote.

\newcommand{\mypar}[1]{\vspace{1mm}\noindent {\bf #1}~~}
\title{Cross-Sensor Touch Generation}
\author{
\textbf{Samanta Rodriguez}\equalcontrib\textsuperscript{1}\quad
\textbf{Yiming Dou}\equalcontrib\textsuperscript{1,2}\quad
\textbf{Miquel Oller}\textsuperscript{1}\quad
\textbf{Andrew Owens}\textsuperscript{1,2}\quad
\textbf{Nima Fazeli}\textsuperscript{1}
\\[0.25em]
\textsuperscript{1}University of Michigan\qquad
\textsuperscript{2}Cornell University
\\[0.25em]
\equalcontrib Equal contribution
}

\begin{document}
\maketitle

%===============================================================================

\begin{abstract}

    Today's visuo-tactile sensors come in many shapes and sizes, making it challenging to develop general-purpose tactile representations. This is because most models are tied to a specific sensor design. To address this challenge, we propose two approaches to cross-sensor image generation. The first is an end-to-end method that leverages paired data (Touch2Touch). The second method builds an intermediate depth representation and does not require paired data (T2D2: Touch-to-Depth-to-Touch). 
    Both methods enable the use of sensor-specific models across multiple sensors via the cross-sensor touch generation process.
    Together, these models offer flexible solutions for sensor translation, depending on data availability and application needs. We demonstrate their effectiveness on downstream tasks such as in-hand pose estimation and behavior cloning, successfully transferring models trained on one sensor to another. Project page: \url{https://samantabelen.github.io/cross_sensor_touch_generation}.
    
    %We demonstrate their effectiveness on downstream tasks such as cup stacking and tool insertion, where models originally designed for one sensor are successfully transferred to another using in-hand pose estimation.
\end{abstract}

\keywords{Tactile Sensing, Manipulation, Representation Learning} 

%===============================================================================

\section{Introduction}
\vspace{-5pt}
\label{introduction}

Tactile sensing is a fundamental enabling technology for dexterous manipulation. Yet, in comparison to their visual counterparts, touch sensors remain highly diverse and lack standardization. 
For example, the robotics community has demonstrated numerous manipulation capabilities \cite{oller2023manipulation, kim2022active, suresh2023midastouch} using a variety of vision-based tactile sensors such as GelSight~\cite{yuan2017gelsight}, Soft Bubble \cite{softbub_tedrake}, GelSlim \cite{gelsim_donlon}, Finger Vision \cite{fingervision}, DIGIT \cite{lambeta2020digit}, and DenseTact \cite{Do2022DenseTactOT}. This sensor diversity poses a significant challenge: specialized algorithms must typically be developed and optimized per sensor. These sensor-specific algorithms are difficult to reuse when their corresponding sensors are unavailable, and adapting them to other sensors can be time-consuming and expensive. Moreover, machine learning models trained on one tactile sensor often fail to generalize to other sensors due to significant distribution shifts.

\begin{figure}[h]
    \centering
    \includegraphics[width=1\linewidth]{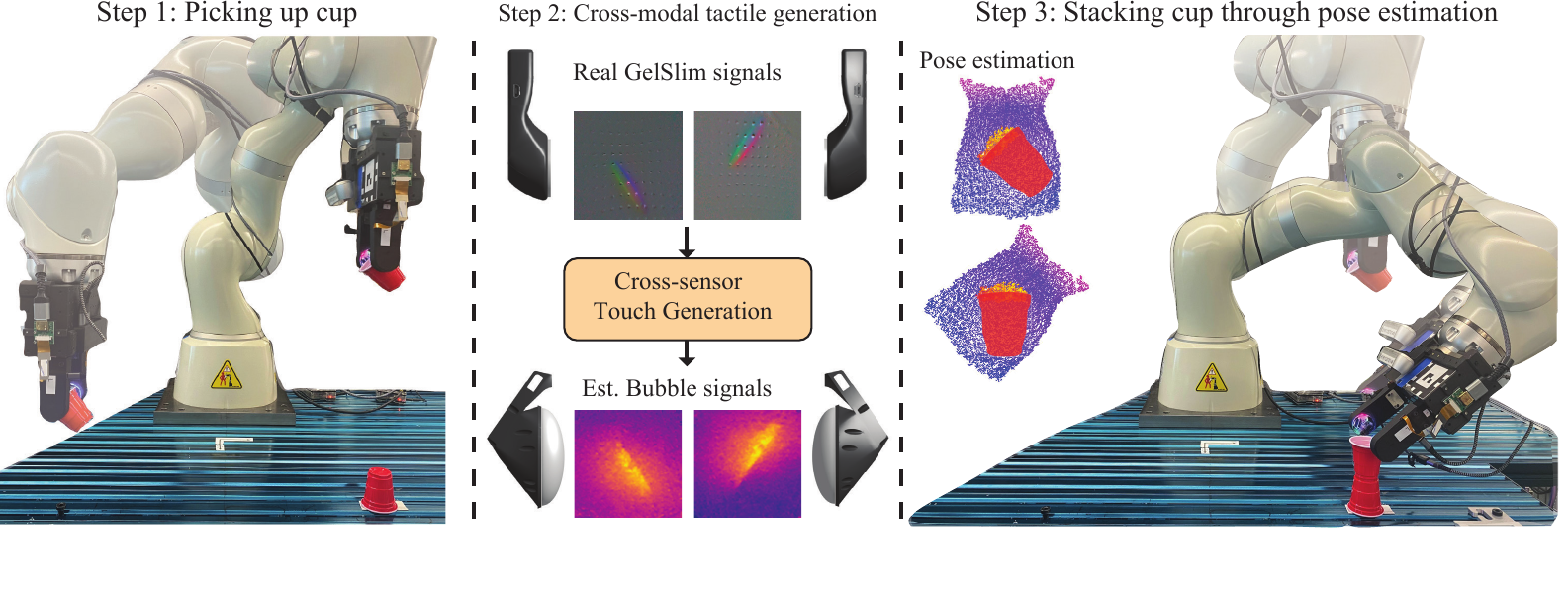}
    \vspace{-8mm}
     \caption{{\bf Transferring manipulation skills between touch sensors via cross-modal prediction.} We execute a manipulation skill designed for one touch sensor (Soft Bubble) on a robot equipped with a different sensor (GelSlim). We demonstrate two approaches to the translation of one touch signal to another --- that is, we predict what the object would have felt like if it were manipulated with Soft Bubble rather than GelSlim. The signal is then used for the downstream skill.} \vspace{-4mm}

    \label{fig:teaser}
\end{figure}

Despite their diversity, vision-based tactile signatures are largely composed of fine-grained shape features~\cite{johnson2009retrographic,yuan2017gelsight,gelsim_donlon,fingervision,lambeta2020digit,Do2022DenseTactOT}. As a result, they convey similar information, such as an object's surface geometry and the contact shape. In this paper, we ask whether this overlapping information can be used to translate tactile signals between sensors, thereby enabling models designed for one tactile sensor to be transferred to another.

We propose two approaches to address this challenge. First, we frame cross-sensor translation as a cross-modal prediction task and train a diffusion model to generate the tactile signal of one sensor conditioned on that of another. We train this model using paired tactile data collected by probing the same object location with two different sensors. Second, we introduce an approach that uses shape as an intermediate representation: we first predict a depth map from the source touch signal, then translate that depth into the target sensor’s signal. This method does not require paired data and leverages the fact that shape is shared across vision-based tactile signals, thereby facilitating the integration of new sensors into the framework with minimal data collection.

% Specifically, we formulate this as a two-stage cross-modal prediction task. First, we train a depth estimation model to predict depth from tactile images. Next, we use the estimated depth to generate the tactile signals as perceived by a different sensor using a generative image-to-image model. To demonstrate the efficacy of our method, we perform cross-modal tactile generation among multiple sensor pairs, including GelSlim, Soft Bubble, and DIGIT sensors.

% AO: I'll leave Sam, Yiming, and Miquel to rewrite the following text. I think this should also be longer (the current intro is short!)
% We evaluate our approach on two manipulation tasks: in-hand object pose estimation and behavior cloning. In both tasks, the robot has access to measurements from the equipped (source) sensor and algorithms for a different (target) sensor. % For example, a robot equipped with one tactile sensor manipulates objects, and our method converts these tactile signals to a different sensor’s modality, allowing the robot to perform tasks using models trained exclusively on the target sensor. 
% The successful execution of these tasks requires an accurate target tactile image and depth generation. 

% We evaluate our approach using in-hand object pose estimation, both as a quantitative metric and as a key component in downstream robotic tasks such as cup stacking and tool insertion (Fig. \ref{fig:teaser}). 
We evaluate our approach on two manipulation tasks: in-hand object pose estimation (Fig. \ref{fig:teaser}) and behavior cloning.
In both tasks, the robot has access to measurements from the equipped (source) sensor and uses algorithms designed for a different (target) sensor. Successful execution requires accurate generation of both the target tactile image and the corresponding depth map.

% Our contributions are threefold; First, we present two approached to enable cross-sensor tactile generation when paired data is available and without the need of paired data; second, we leverage existing visio-tactile image datasets to enhance the model training and generalization; and third, we present a modular and flexible framework that enables seamlessly integration of new sensors, facilitating the generation of tactile data across various sensor modalities. 

Our contributions are as follows:
First, we present two approaches for cross-sensor tactile signal generation: one that uses paired data with a conditional diffusion model to produce fine-grained target signals, and another that leverages depth as an intermediate representation, enabling signal translation without the need for paired data.
% Second, we demonstrate the flexibility of our modular framework by adapting to novel sensors using unpaired data, and by leveraging foundation models—such as Depth Anything \cite{depth_anything_v2} —for depth estimation, reducing data collection requirements.
Second, we evaluate cross-sensor transfer with quantitative and qualitative image generation metrics, and in-hand object pose estimation as a tactile-specific metric. 
Third, we demonstrate the practical utility of our approach in robotic manipulation scenarios, showing that it enables precise task transfer across heterogeneous tactile sensors.

\vspace{-8pt}
\section{Related Work}
\vspace{-8pt}
\label{related_work}

% \mypar{Vision-based tactile sensing.}
% Sensor

\noindent\textbf{Vision-based tactile sensing.} In the last decade, the robotics community has adopted a variety of vision-based tactile sensors, such as GelSight~\cite{yuan2017gelsight,johnson2009retrographic}, Soft Bubble~\cite{softbub_tedrake}, GelSlim~\cite{gelsim_donlon}, Finger Vision~\cite{fingervision}, DIGIT~\cite{lambeta2020digit}, and  DenseTact~\cite{Do2022DenseTactOT}. These sensors convert touch signals into vision-like signals, representing touch as 2D images or 3D representations (e.g., point clouds). These sensors are rapidly gaining popularity and have proven valuable in a variety of applications~\cite{calandra2018more,digit_tactile_sensor,oller2023manipulation,li2014localization,kim2022active}. We use the Soft Bubble~\cite{kuppuswamy2019fast}, DIGIT~\cite{suresh2023midastouch} and GelSlim~\cite{taylor2022gelslim} sensors in our experiments. The Soft Bubble~\cite{kuppuswamy2019fast} is composed of a thin, highly compliant, air-filled membrane paired with a camera-based depth sensor. Tactile signatures are perceived as deformations of the membrane due to external contacts. The DIGIT~\cite{suresh2023midastouch} and GelSlim~\cite{taylor2022gelslim} measure deformations of an elastomeric skin illuminated by multi-colored LEDs using an RGB camera. We choose these three sensors because of their vastly different deformations and compliance, contact areas, image quality, and 3D (vs. 2D) representation.
As for the algorithms, existing manipulation, perception and controls representations are tied to specific touch sensors. For example, a variety of methods leverage sensor specific local geometry, contact force estimation, or texture~\cite{yang2023generating,li2014localization}. Further, in-hand object pose estimation algorithms have been developed for different visuotactile sensors (e.g., for Soft Bubble~\cite{kuppuswamy2019fast}, GelSlim~\cite{kim2022active}, and DIGIT~\cite{suresh2023midastouch}), for local geometry estimation (e.g., Soft Bubble~\cite{kuppuswamy2019fast}, GelSlim~\cite{taylor2022gelslim}, and DIGIT~\cite{xu2023visual}).
% for force field estimation across the contact area (e.g., Soft Bubble~\cite{kuppuswamy2020soft}, GelSlim~\cite{taylor2022gelslim}, Finger Vision~\cite{yamaguchi2016combining}). 
We reduce the need for sensor-specific methods by enabling models to transform one touch signal to another. 

\noindent\textbf{Cross-sensor tactile representation.} Other work aims to design a unified representation for tactile sensors. UniTouch~\cite{yang2024binding} and AnyTouch~\cite{feng2025anytouch} align the tactile embeddings with pretrained image and text embeddings, and supports multi-sensor training by sensor-specific tokens. Sparsh~\cite{higuera2024sparsh} and T3~\cite{t3} encode images from different sensors into a shared representation space and train the encoders with self-supervised learning. CTTP~\cite{rodriguez2024contrastive} uses paired tactile data to learn a touch feature space based on contrastive learning. This method requires all downstream methods to operate on a special feature representation.
In contrast, we focus on the explicit transfer of raw touch signals, without the need for a latent feature space. This brings two major advantages: (i) we can use it for tasks that require explicit geometry from the raw tactile signals, and (ii) we can apply it to existing downstream touch processing models ``zero shot'', without adaptation. As an example, we directly apply iterative closest point (ICP) on a generated Soft Bubble image to perform robotic object manipulation tasks (Sec.~\ref{experiments}). 

%Our method tackles with the problem by transferring across different tactile signals explicitly in the image space. 
%In contrast to prior works that learn representations in the latent space, our method  fine-grained geometric details from the touch images, e.g., cup stacking with ICP (Sec.~\ref{experiments}). 
%In contrast to prior works that learn representations in the latent space, our method can be applied to manipulation tasks that require fine-grained geometric details from the touch images, e.g., cup stacking with ICP (Sec.~\ref{experiments}). 

% \mypar{}
\noindent\textbf{Cross-modal generation.} A variety of early generative models transformed images from one format to another~\cite{isola2017image, sangkloy2017scribbler, hu2017toward, wang2018high}.
Recent works in cross-modal image translation frequently use diffusion~\cite{sohl2015deep} for its ability to generate high-quality images with stable training. These models have been used with a variety of different conditioning signals, resulting in models that perform text-to-image~\cite{avrahami2022blended,kawar2023imagic,nichol2021glide,saharia2022photorealistic,rombach2022high,zhang2023adding,ruiz2023dreambooth}, audio-to-image~\cite{girdhar2023imagebind}, video-to-audio~\cite{luo2024diff}, etc. % text-to-3D~\cite{luo2021diffusion,shao2022diffustereo,poole2022dreamfusion,lin2023magic3d}, % these are based on score distillation sampling, which seems very different
Our work is closely related to methods that estimate touch from vision. These works have proposed models under various settings, including desktop~\cite{li2019connecting}, object-centric~\cite{gao2023objectfolder}, sub-scene~\cite{yang2022touch,yang2024binding}, and full-scene~\cite{dou2024tactile}. Like many of these works~\cite{higuera2023learning,yang2023generating,yang2024binding,dou2024tactile, caddeo2024sim2real}, we use diffusion to generate touch signals. 
However, our conditioning is based on the touch signal from another sensor rather than from a visual signal. Our framework transforms across significantly different visuo-tactile sensors, including Soft Bubble (not gel-based), GelSlim, and DIGIT (gel-based). To the best of our knowledge, this work is the first to address cross-sensor generation.

% , and demonstrates improved performance in downstream tasks.

% \input{text/dataset}
\vspace{-5pt}
\section{Method}
\vspace{-5pt}

\label{methodology}
\begin{figure}[b]
    \centering
    \includegraphics[width=1.0\linewidth]{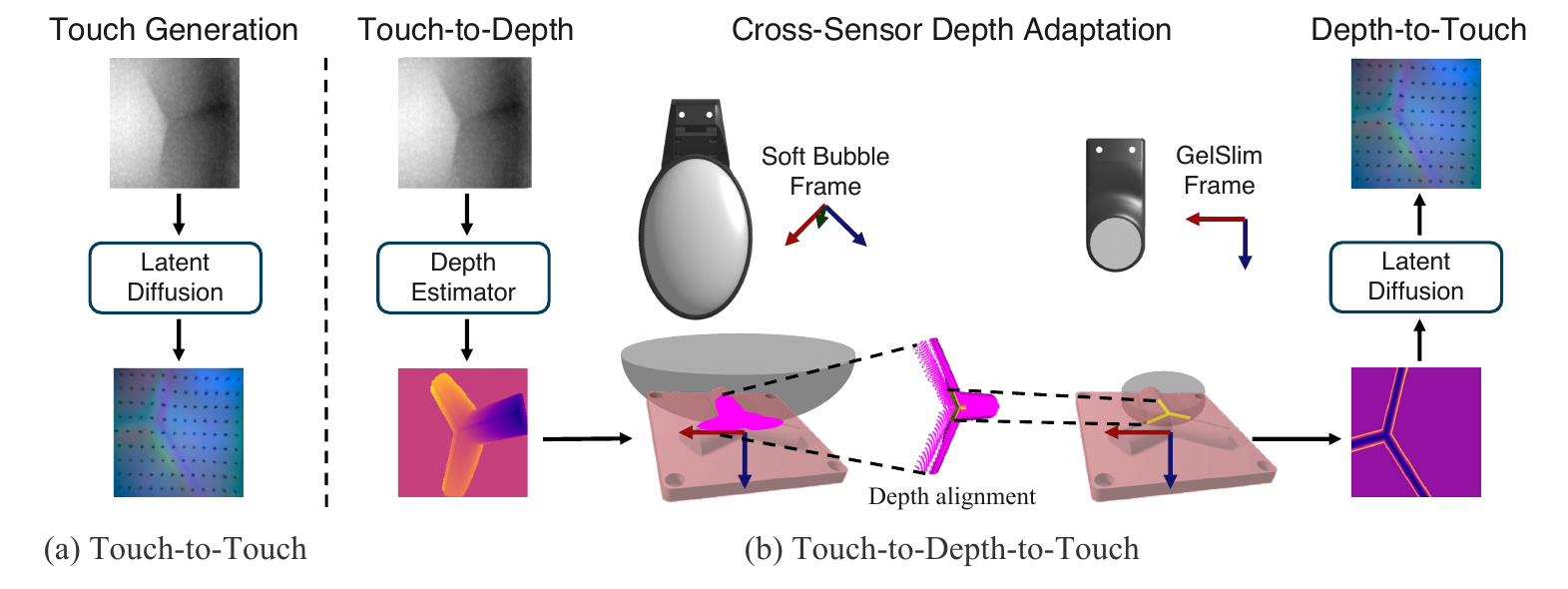}
    \caption{{\bf Translating signals between touch sensors.} We investigate two different approaches to cross-sensor touch translation. (a) We train a latent diffusion model to direct predict one sensor's signal from another's, using paired training data. (b) We use depth as an intermediate representation, thus avoiding the need for paired training data. We predict depth from touch, adapt the depth map to match the specifications of another sensor, then generate a touch signal from the resulting depth map. We use the resulting touch translation models for robotic manipulation tasks.}
    \label{fig:method}
\end{figure}
In this section, we first describe our data collection processes. Next, we propose two approaches (shown in Fig. \ref{fig:method}) for cross-sensor tactile generation: (1) one-stage end-to-end generation using paired touch signals, and (2) two-stage generation by using depth as an intermediary representation.

\vspace{-5pt}
\subsection{Dataset}
\vspace{-5pt}
\label{dataset}
%In the following, we describe the two main data collection processes and data statistics.

We use a robot to collect paired touch data, such that two different sensors probe the same physical location. We also obtain touch paired with depth data, which we use for a model that predicts depth as an intermediate representation for touch translation.

\noindent\textbf{Collecting Paired Multimodal Tactile Signals.}
We use a KUKA LBR Med 14 R820 robotic arm with a WSG-50 gripper to collect paired tactile signals from two distinct sensors: Soft Bubble and GelSlim. These sensors differ significantly in mechanical design, contact compliance, contact area, and spatial resolution, necessitating a highly repeatable and spatially precise setup for generating semantically aligned tactile pairs.
To ensure alignment and both sensors capture meaningful geometric features under consistent conditions the gripper is positioned so that the center of each sensor contacts the same point on the target object, and compliance differences are compensated by adjusting the gripper.
Paired signals must share \textit{mutual semantic information}—that is, both must capture recognizable features despite differences in spatial resolution and contact area. For example, the Soft Bubble sensor covers roughly $16\times$ more surface area than the GelSlim but at a much lower resolution (2.36 vs. 23.72 pixels/mm). To address this, we carefully select objects with distinctive geometry (e.g., the elbow of a hex key) and post-process the data to ensure that both signals contain salient, recognizable features within their respective sensing capabilities.

% pairs contain semantically meaningful and distinctive features of each object (e.g., the elbow of a hex key). 

% preserve key features of the tactile signatures across sensors by ensuring that all the touch signals in our dataset keep the distinctive features of each object (e.g., the elbow of a hex key) and by selecting objects that possess distinct features visible with the resolution of each sensor.

% It is important that key features of objects are present in both sensor images to address the significant difference in contact area and tactile signal resolution between the Soft Bubble and GelSlim sensors. On one hand, the Soft Bubble sensor covers approximately $16 \times$ the contact area  of the GelSlim sensor, as shown in Fig.~\ref{fig:touch_signals}. On the other hand, the GelSlim sensor signals are much more detailed than the Soft Bubble sensor signals: they provide 23.72 pixels/mm versus 2.36 pixels/mm for the Soft Bubble. This difference in resolution allows the GelSlim to render precise microgeometry in the contact area.

\noindent\textbf{A Dataset of Multimodal Touch-Depth Pairs.}
To enable supervised learning for both tactile-to-depth and depth-to-tactile translation, we construct a dataset of paired tactile and depth observations following a procedure inspired by \cite{derendering}. Data is collected using a robot equipped with a variety of vision-based tactile sensors, including Soft Bubbles, GelSlims, DIGITs, and dotless GelSlim variants, offering diverse tactile imaging characteristics.
We use 12 indenters of known geometry, spanning flat, curved, and angled profiles. Each indenter is pressed into the sensor surface within a controlled 3D sampling grid covering $10,\mathrm{mm} \times 10,\mathrm{mm}$ in translation and up to $45^\circ$ in orientation, centered at the indenter's origin. This ensures uniform spatial coverage and varied contact patterns. For each indenter, we collect 60 tactile samples at distinct poses, resulting in a total of 720 samples per sensor.
Each tactile observation is paired with a ground-truth depth map that captures the 3D contact surface. These depth maps are generated using the robot’s proprioceptive data, calibrated camera models, and geometric rendering based on the known object and contact poses.

% We propose T2D2, a framework for cross-modal tactile generation that translates tactile images from one visuo-tactile sensor to those of other sensors by using depth as an intermediary representation. Our approach is structured in three stages: (1) a depth estimation model that extracts an intermediate depth map from the source tactile image; (2) a depth adaptation stage that aligns this depth information to the target sensor's domain; and (3) a tactile image generation model that synthesizes the tactile image of the target sensor based on the adapted depth map.

\vspace{-3pt}
\subsection{One-Stage Generation using Paired Touch Signals}
\vspace{-5pt}
\label{sec:t2t}
For the one-stage approach, we train a cross-modal diffusion model to directly translate from one sensor to another based on paired tactile images. We refer to this method as touch-to-touch (T2T).
For the implementation, we use a generative model based on latent diffusion~\cite{rombach2022high} to generate the touch signal of the target sensor by conditioning on the source measurement (Fig.~\ref{fig:method}). We use a ResNet-18~\cite{he2016deep} to encode the touch images from the source sensor into a 2D feature map. The feature map is then concatenated with the noise signal and passed into the denoising UNet. 

We specifically apply our model to translating from the GelSlim to Soft Bubble sensors (and vice versa). When the Soft Bubble sensor is used as the target sensor, we inflate the 1-channel signals into 3-channel signals by tiling the image channel-wise.  When the Soft Bubble tactile signal is generated from the GelSlim image using diffusion, we perform three post-processing steps to ensure accurate tactile information generation (beyond visual fidelity). First, we take the average value of the three channels of the prediction to map it back to one channel. Next, we normalize the prediction (in the range $[-1, 1]$) back to the depth map values by using the maximum and minimum values of the depth maps across the training dataset. Finally, to deal with small scaling/bias in the predictions (which can drastically change their interpretations as point clouds), we shift the pixel values of the generated Soft Bubble images. We calculate the mean and standard deviation of the generated and ground truth Soft Bubble images on the training dataset and use these values to renormalize the generated images.

In our experiments, we compare our end-to-end cross-sensor generation model to VQ-VAE, a commonly used baseline for cross-modal generation. We provide implementation details in the supplementary material.

% \begin{wrapfigure}[10]{R}{0.5\textwidth}
% \vspace{-3mm}
%  \center{
%    \includegraphics[width=0.98\linewidth]{figures/t2t_ldm.pdf}\vspace{-2mm}
%    \captionof{figure}{{\bf Cross-sensor generation.} We train a latent diffusion model to directly generate one touch sensor's signal from another. For example, we predict Soft Bubble signal from GelSlim, two sensors with very different size and compliance.} \label{fig:t2tldm}}
% \end{wrapfigure}

\vspace{-3pt}
\subsection{Two-Stage Generation using Unpaired Touch Signals}
\vspace{-3pt}

The one-stage method can be trained end-to-end, but it requires paired touch signals, which can be difficult to collect in practice. Therefore, we also propose a two-stage framework that translates touch to a depth map as an intermediate representation. This ``touch-to-depth-to-touch'' (T2D2) approach enables training without paired touch signals. This model contains three modules: (i) a depth estimation model that predicts a depth map from the source tactile image, (ii) a depth adaptation stage that modifies the source sensor depth map to match the specifications of target sensor and (iii) a tactile image generation model that synthesizes the target sensor output based on the adapted depth map.

%\subsubsection{Depth Estimation Model}
\mypar{Depth estimation model.} Our goal is to estimate a depth map from a touch signal. This requires capturing both the contact geometry of the source sensor and the geometry of the object surface. We adapt Depth Anything V2~\citep{depth_anything_v2}, a state-of-the-art monocular depth estimation model, to this task. 

We modify the model to jointly estimate both the depth map and a contact mask, which indicates the sensor parts that are in contact with the object. To do this, we add a decoder head that takes intermediate high-resolution features from the depth decoder to predict a binary contact map. 
We train the model on real touch signals paired with ground-truth depth maps, which are obtained from a neural tactile de-rendering method~\citep{derendering} (Sec.~\ref{dataset}).

%\subsubsection{Depth Adaptation}
%\subsubsection{Obtaining Sensor-specific Depth Maps}
\mypar{Depth adaptation.}
Before converting a given depth map into a touch signal, we need to determine which parts of it are visible (e.g., due to a sensor's limited field of view) and to transform it into the coordinate system of the sensor (e.g., accounting for the fact the pose of the vision-based sensor's camera). We call this process {\em depth adaptation}. For the adaptation stage, we use the depth map \(D_S'\) and its corresponding binary mask \(M_S'\) obtained from our depth estimation model to find the equivalent sensor-specific depth map \(D_T''\) and mask \(M_T''\) as if the target sensor was in contact. 

%from the source sensor tactile image \(I_S\)
% Mention that we also need the K_T instrinsic matrix and M_T (mesh) geometry of target sensor and T_S->T (Add alignment frame explanation here and) 

% First, we define the set of valid pixel coordinates, using the contact mask \(M_S'\), as follows:
% \begin{equation}
% \Omega = \{ (u,v) \mid M_S'(u,v) = 1 \}
% \end{equation}

To perform this adaptation, we first determine the set of pixels $\Omega$ that are indicated by the predicted mask $M_S'$ to be in contact with the sensor.
For each such pixel \((u,v) \in \Omega\), we transform it as follows. We back-project the depth value into 3D space using the vision-based touch sensor's camera intrinsic matrix \(K_S^{-1}\) and transform the resulting 3D point from the source sensor frame to the target sensor frame using the rigid transformation \(T_{S \rightarrow T}\). Specifically, \(T_{S \rightarrow T}\) is defined as the composition of the transformation from the source sensor to the alignment frame, \(T_{S \rightarrow A}\), and from the alignment frame to the target sensor, \(T_{A \rightarrow T}\); that is, \(T_{S \rightarrow T} = T_{A \rightarrow T} \circ T_{S \rightarrow A}\). The alignment frame is defined as a common reference frame shared across different tactile sensors. This standardization allows for consistent interpretation of the contact geometry, regardless of each sensor's unique geometry or pose. This process is expressed as:
\begin{equation}
    \mathcal{P}_T = \left\{ T_{S \rightarrow T} \left( D_S'(u,v) \cdot K_S^{-1} \begin{pmatrix} u \\ v \\ 1 \end{pmatrix} \right) : (u,v) \in \Omega \right\},
    \label{eq:estimated_target_pcd}
\end{equation}
where \(\mathcal{P}_T\) represents the resulting point cloud in the target sensor frame. We then use $\mathcal{P}_T$ to obtain the target sensor's depth map and its contact mask. We provide details in the supplementary material.

\paragraph{Generating touch from depth.} 
Given a sensor-specific depth map, we generate a corresponding touch signal using diffusion. To do this, we use a model very similar to the one described in Sec.~\ref{sec:t2t}, but with depth conditioning in lieu of touch conditioning. 

%\subsubsection{Tactile Image Generation Model}
%The image generation model of follows a similar structure to T2T, with the only difference in the conditioning module: since T2D2 uses depth as the intermediary representation, the conditioning module takes the adapted depth map instead of the touch signal from the source sensor as input. 

\vspace{-3pt}
\section{Experiments}
\vspace{-3pt}
\label{experiments}
We evaluate cross-sensor touch generation using visual metrics, a tactile-specific metric, and downstream robotic tasks.
\vspace{-8pt}
\subsection{Evaluation Metrics}
\vspace{-8pt}
\paragraph{Visual Metrics.}
The visual quality of the generated tactile images is assessed with three standard image-based metrics: Peak Signal-to-Noise Ratio (PSNR), Structural Similarity Index Measure (SSIM), and Fréchet Inception Distance (FID). PSNR and SSIM measure pixel-level fidelity and perceptual similarity to the ground truth, respectively, while FID quantifies distributional alignment between generated and real samples.
\vspace{-14pt}
\paragraph{Tactile-Specific Metric.}
% For our tactile-specific metric, we utilize in-hand object pose estimation, where a model is trained to estimate in-hand object pose on the target sensor tactile images and applied to the source sensor without retraining or modification, leveraging our method to translate the source tactile images. We report translation errors and angular errors in pose estimation as our metric.
As our tactile-specific metric, we use in-hand object pose estimation, where a neural network trained on the target sensor tactile images is applied to source sensor images translated by our method. Performance is quantified by translation and angular pose errors. 

\begin{figure}[h!]
    \centering
    \includegraphics[width=\columnwidth]{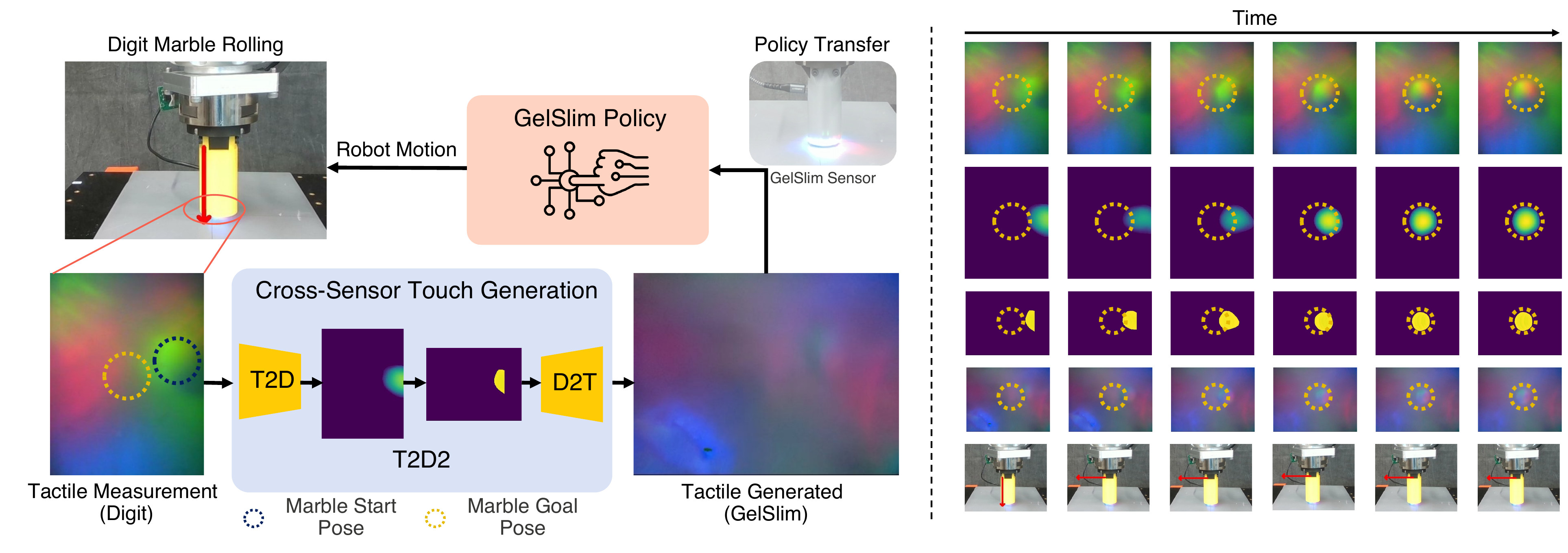}
    \vspace{-5mm}
    \caption{\textbf{Marble rolling policy transfer via T2D2.} We train behavior-cloning policy on GelSlim tactile images to roll a marble from random starts to the image center. At test time on DIGIT, we translate each DIGIT tactile signature to its GelSlim counterpart with T2D2 and run the same policy unchanged. \textbf{Left:} pipeline—DIGIT → (T2D2) → GelSlim → GelSlim-trained policy. \textbf{Right:} transferred roll-outs on DIGIT converge to the center (zero-shot; no retraining).}
    \label{fig:real_experiments}
    \vspace{-1mm}
\end{figure}

\vspace{-14pt}
\paragraph{Downstream Robotic Tasks.}
We further evaluate cross-sensor touch generation on two downstream robotic tasks: peg-in-hole insertion and marble rolling. We perform the peg-in-hole insertion task, originally designed for the Soft Bubble sensor, using a robot equipped with GelSlim sensors. Our T2T model predicts the corresponding Soft Bubble signal from the GelSlim reading, which is then used to estimate object pose via Iterative Closest Point (ICP) alignment. This pose estimate enables the robot to complete tool insertion and cup stacking with previously unseen objects, as illustrated in Fig.~\ref{fig:teaser}. For the marble rolling task, we train a behavior cloning policy to roll a marble—initially placed at a random position—toward the center of the GelSlim sensor image. The policy is trained using supervised learning on GelSlim tactile data and learns to guide the marble using the tactile feedback provided by the sensor. We then test the transferability of this policy to a different sensor, DIGIT, by using our T2D2 model to generate the corresponding GelSlim tactile signal from the DIGIT input. This allows the original policy, trained exclusively on GelSlim data, to be applied to DIGIT inputs without any retraining or modification. This process is shown in Fig. \ref{fig:real_experiments}.

\begin{figure*}
    \centering
    \includegraphics[width=1\linewidth]{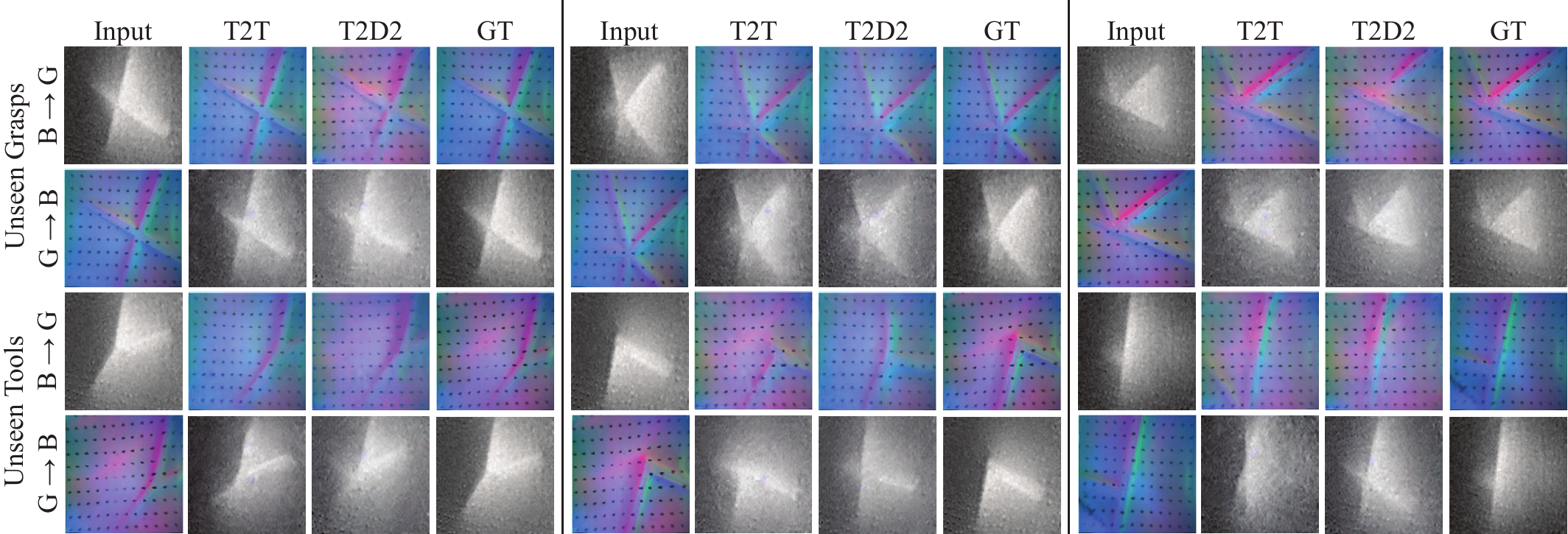}
    \caption{\textbf{Generation Qualitative Results}. Qualitative results for unseen grasps and tools using T2T and T2D2. Rows indicate sensor transfer directions (B → G, G → B); columns show input, model outputs, and ground truth.}
    \vspace{-15pt}
    \label{fig:t2t_vs_t2d2_qualitative_results}
\end{figure*}

\begin{figure}
    \begin{flushright}
        \includegraphics[width=0.98\linewidth]{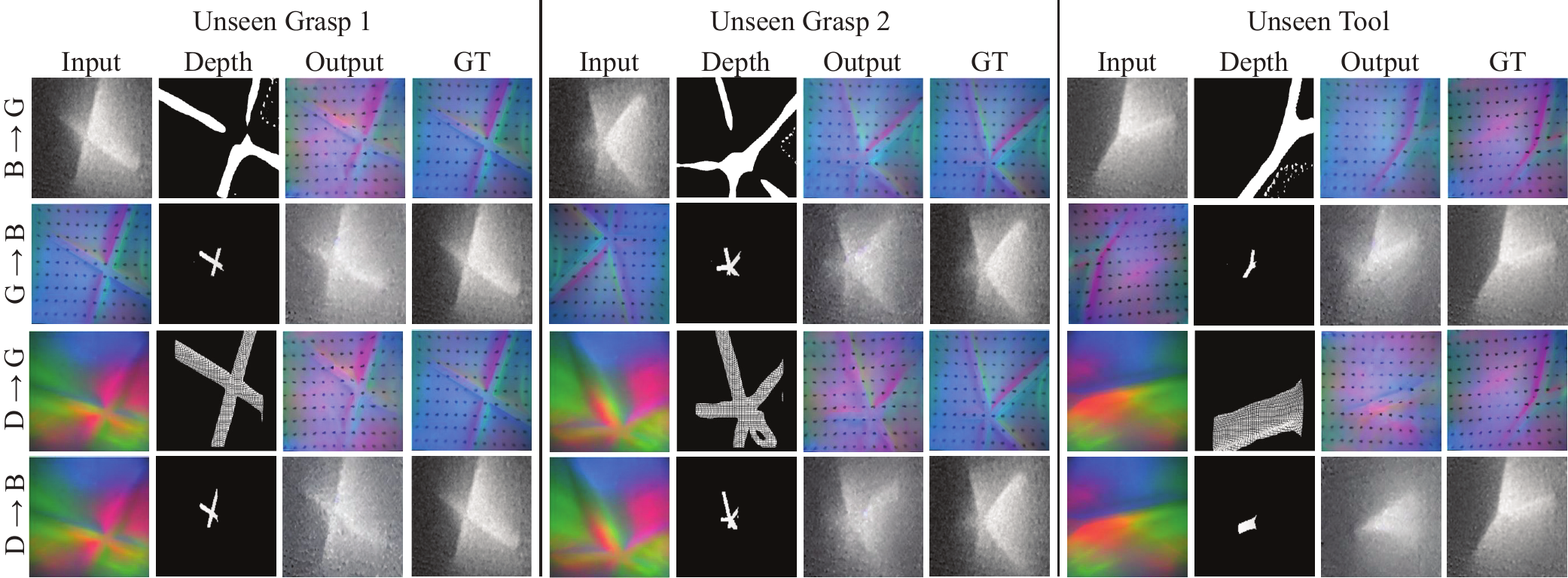}
        \caption{\textbf{T2D2 Qualitative Results}. Evaluation of T2D2 model on unseen grasps and tools. Each block shows the input, adapted depth map, generated tactile output, and ground truth (GT) for various sensor transfers.}
        \vspace{-15pt}
        \label{fig:t2d2_depth_qualitative_results}
    \end{flushright}
\end{figure}

\subsection{Visual and Tactile Metrics Results}
The results in Table~\ref{tab:evaluation_metrics} and Table~\ref{tab:evaluation_metrics_unseen_tools} show that while T2D2 enables cross-sensor tactile generation using unpaired data and a depth-based intermediate representation, this flexibility comes at the cost of fine-grained accuracy necessary for downstream tasks like pose estimation. In contrast, T2T, which performs direct image-to-image translation, preserves more structural fidelity, resulting in lower translation and angular errors. Fig. \ref{fig:t2t_vs_t2d2_qualitative_results} shows qualitative results for both methods.

This trend is also reflected in the visual metrics. T2T consistently achieves higher PSNR, SSIM, and lower FID scores than T2D2, indicating sharper, more perceptually accurate, and distributionally faithful outputs. For example, in the Bubbles to GelSlims transfer on unseen grasps, T2T achieves a PSNR of 30.92 and SSIM of 0.93, compared to 21.69 and 0.83 for T2D2. Similar trends are observed across unseen tools as well.

While PSNR values are relatively comparable between the GelSlims to Bubbles and Bubbles to GelSlims translations, all other metrics indicate that translating from GelSlims to Bubbles is a more challenging task. This is especially evident in the angular and translation errors, which are consistently higher in the GelSlims to Bubbles direction for both models. A key reason is that the Bubble sensor is considerably larger than the GelSlim sensor, requiring the models to effectively learn to outpaint or infer tactile signals beyond the spatial extent of the input. In contrast, translating from Bubble to GelSlim remains within the bounds of the original signal, making the task less ambiguous.

T2D2 also shows an increased error when transferring from Digits to GelSlims and from Digits to Bubbles, reflecting the added challenge of translating from a structurally different sensor. Still, qualitative results in both directions are strong, indicating that the generated signals retain coherent tactile features, 
shown in Fig. \ref{fig:t2d2_depth_qualitative_results}. 
Incorporating Digits into the T2D2 pipeline was efficient, as the model supports unpaired training, and fine-tuning the depth estimator required collecting only about one-tenth of the paired data typically needed to train an end-to-end model. These results demonstrate the scalability of T2D2 for incorporating new sensors with minimal supervision.
\vspace{-6pt}
\begin{table}[htbp]
\centering
\small
\begin{tabularx}{\textwidth}{l l|Y{0.7cm} Y{0.7cm} Y{0.9cm}|c|c}
\toprule
\textbf{Transfer} & \textbf{Model} & \textbf{PSNR} & \textbf{SSIM} & \textbf{FID} & \textbf{Trans. Error [mm]} & \textbf{$\theta$ Error [$^{\circ}$]} \\
&& $\uparrow$ & $\uparrow$ & $\downarrow$ & $\downarrow$ & $\downarrow$ \\
\midrule
Bubbles GT & - & - & - & - & $0.22 \pm 0.12$ & $0.51 \pm 0.41$ \\
GelSlims $\rightarrow$ Bubbles & T2T & 25.85 & 0.59 & 135.66 & $0.92 \pm 0.58$ & $1.87 \pm 1.23$ \\
GelSlims $\rightarrow$ Bubbles & T2D2 & 20.51 & 0.52 & 122.06 & $3.23 \pm 1.73$ & $11.53 \pm 9.09$ \\
Digits $\rightarrow$ Bubbles & T2D2 & 20.53 & 0.48 & 111.77 & $3.38 \pm 1.86$ & $9.85 \pm 8.10$ \\
\midrule
GelSlims GT & - & - & - & - & $0.29 \pm 0.17$ & $0.83 \pm 0.59$ \\
Bubbles $\rightarrow$ GelSlims & T2T & 30.92 & 0.93 & 33.25 & $0.40 \pm 0.29$ & $1.27 \pm 0.97$ \\
Bubbles $\rightarrow$ GelSlims & T2D2 & 21.69 & 0.83 & 53.99 & $2.63 \pm 1.93$ & $8.47 \pm 6.82$ \\
Digits $\rightarrow$ GelSlims & T2D2 & 17.73 & 0.72 & 65.97 & $4.10 \pm 1.93$ & $17.67 \pm 11.47$ \\
\bottomrule
\end{tabularx}
\vspace{3pt}
\caption{\textbf{Unseen Grasps}. Evaluation metrics for cross-modal tactile generation tasks, including visual (PSNR, SSIM, FID) and tactile-specific (translation and angular error) metrics.}
\label{tab:evaluation_metrics}
\end{table}
\vspace{-20pt}

\begin{table}[htbp]
\centering
\small
\begin{tabularx}{\textwidth}{l l|Y{0.7cm} Y{0.7cm} Y{0.9cm}|c|c}
\toprule
\textbf{Transfer} & \textbf{Model} & \textbf{PSNR} & \textbf{SSIM} & \textbf{FID} & \textbf{Trans. Error [mm]} & \textbf{$\theta$ Error [$^{\circ}$]} \\
&& $\uparrow$ & $\uparrow$ & $\downarrow$ & $\downarrow$ & $\downarrow$ \\
\midrule
Bubbles GT & - & - & - & - & $0.42 \pm 0.26$ & $1.03 \pm 0.90$ \\
GelSlims $\rightarrow$ Bubbles & T2T & 21.77 & 0.48 & 167.72 & $3.40 \pm 1.82$ & $14.00 \pm 8.13$ \\
GelSlims $\rightarrow$ Bubbles & T2D2 & 18.70 & 0.45 & 134.74 & $4.40 \pm 2.14$ & $12.71 \pm 9.57$ \\
Digits $\rightarrow$ Bubbles & T2D2 & 18.91 & 0.46 & 129.30 & $3.63 \pm 2.29$ & $14.75 \pm 13.83$ \\
\midrule
GelSlims GT & - & - & - & - & $0.78 \pm 0.52$ & $1.71 \pm 1.95$ \\
Bubbles $\rightarrow$ GelSlims & T2T & 21.50 & 0.81 & 62.38 & $2.53 \pm 1.56$ & $6.35 \pm 8.43$ \\
Bubbles $\rightarrow$ GelSlims & T2D2 & 18.33 & 0.73 & 68.67 & $3.58 \pm 1.90$ & $9.99 \pm 8.08$ \\
Digits $\rightarrow$ GelSlims & T2D2 & 17.37 & 0.68 & 105.30 & $4.87 \pm 2.18$ & $13.07 \pm 9.73$ \\
\bottomrule
\end{tabularx}
\vspace{3pt}
\caption{\textbf{Unseen Tools}. Evaluation metrics for cross-modal tactile generation tasks, including visual (PSNR, SSIM, FID) and tactile-specific (translation and angular error) metrics.}
\label{tab:evaluation_metrics_unseen_tools}
\end{table}
\vspace{-10pt}

\subsection{Downstream Robotic Task Results}
\vspace{-4pt}
For the peg-in-hole insertion task, we compare T2T with another end-to-end generative method, VQ-VAE. In Table~\ref{tab:Insertion Task Success}, T2T consistently outperforms VQ-VAE across all tasks, achieving higher success rates in both tool-based (unseen tools from our dataset) and real-object scenarios. The largest improvements are observed in pencil insertion (21/30 vs. 7/30) and tool 1 insertion (18/30 vs. 9/30), demonstrating T2T’s effectiveness in generating accurate cross-modal tactile signals for manipulation. For the marble rolling task, the policy succeeds in \textbf{15 out of 20 trials} when using DIGIT inputs translated to GelSlim, compared to \textbf{20 out of 20} using real GelSlim signals. This experiment demonstrates that our cross-sensor generation framework supports not only perception tasks but also simple sensor-conditioned control, enabling policy reuse across different tactile hardware.
% \vspace{-10pt}

% \vspace{-10pt}
\begin{table}[h!] %\vspace{-0.5cm}
\small
    \centering
    \renewcommand{\arraystretch}{1.1}
    \small
    \vspace{-0.2cm}
    \begin{tabularx}{\columnwidth}{l| c c c |c |c}
        \toprule
        Method & Tool 1 Insertion & Tool 2 Insertion & Tool 3 Insertion & Pencil Insertion & Cup Stacking \\
        \midrule
        VQ-VAE &  9/30 & 8/30 & 15/30 & 7/30 & 21/30 \\
        T2T    & 18/30 & 10/30 & 15/30 & 21/30 & 22/30 \\
        % T2D2   &  19/30 & X/30 & X/30 & X/30 & X/30 \\
        \bottomrule
    \end{tabularx}
    \vspace{3pt}
    \caption{Insertion and Cup Stacking Tasks Success on Unseen Objects}
    \label{tab:Insertion Task Success}
\end{table}
% \vspace{-14pt}

% \vspace{-10pt}
\section{Conclusion}
% \vspace{-8pt}
\label{discussion}

Our exploration of cross-sensor tactile generation reveals several key insights about the nature of visuo-tactile sensing and the utility of generative models in bridging sensor-specific differences. One of the most important observations is that, despite large differences in hardware design and signal modalities, tactile data from different sensors often encodes fundamentally similar geometric and contact information. This latent consistency enables translation between sensor outputs, allowing models trained for one sensor to generalize to others through learned generative mappings.

The performance of the two proposed approaches—T2T and T2D2—also underscores the tradeoff between data requirements and model fidelity. T2T, which leverages paired data, consistently generates high-fidelity, structurally accurate tactile signals, achieving strong performance on both visual and tactile-specific metrics. In contrast, T2D2 offers greater flexibility by requiring only unpaired data and an intermediate depth representation. However, this flexibility introduces challenges in preserving high-frequency geometry and leads to degradation in downstream performance, particularly for precision-sensitive tasks like in-hand pose estimation.

Another important outcome of our work is the demonstration that generative translation between sensors is a viable path to sensor interoperability. Instead of attempting to design universal representations or train all models jointly across multiple sensors, we show that translating touch from one sensor to another can unlock existing, optimized pipelines without any retraining. This capability lowers the barrier to reuse in tactile perception and manipulation systems and opens new directions for modular tactile intelligence, where sensor-specific capabilities can be shared or borrowed on demand via generative translation.
\clearpage
\vspace{-8pt}
\section{Limitations}
\vspace{-8pt}
\label{limitations}

While our study demonstrates the feasibility of cross-sensor tactile generation, several limitations remain. First, the T2T model requires precisely paired tactile data across sensors, which can be time-consuming and hardware-intensive to collect, especially when working with sensors that differ significantly in contact geometry, field of view, or resolution. Although T2D2 addresses this by operating with unpaired data, its reliance on accurate depth estimation and calibration introduces sensitivity to errors in alignment and sensor-specific intrinsic parameters.

Second, both approaches assume that there exists sufficient semantic overlap between the source and target signals—i.e., that the contact patches capture comparable geometric features. This assumption may not hold when sensors have drastically different contact modalities or when the object geometry is highly complex or non-rigid. In such cases, generation fidelity may suffer due to the model having to hallucinate large, unobserved regions, particularly when translating from a smaller to a larger contact area (e.g., GelSlim to Bubble).

Finally, while our models successfully transfer across three sensor types and demonstrate strong downstream task performance, they are limited to vision-based tactile sensors and specific manipulation settings. Extending this framework to other tactile modalities (e.g., resistive, magnetic, or proprioceptive sensors), or to more dynamic or continuous interactions, may require significant architectural adaptations. As tactile sensing systems continue to diversify, generalization across modalities—not just across hardware—remains an open and challenging frontier.

\clearpage
% The acknowledgments are automatically included only in the final and preprint versions of the paper.
\acknowledgments{This work was supported in part by the NSF GRFP (Award No. 2241144), the NSF CAREER program (Award Nos. 2339071 and 2337870), and the NSF NRI (Award No. 2220876). Any opinions, findings, and conclusions or recommendations expressed in this material are those of the authors and do not necessarily reflect the views of the National Science Foundation.}

%===============================================================================

\bibliography{bibliography}  % .bib
\clearpage
\appendix
% \section{Appendix: Behavior Cloning Policy Transfer for Marble Rolling}

% To further evaluate the effectiveness of our cross-sensor tactile generation approach, we conducted an additional experiment on a marble rolling task. In this setup, a behavior cloning policy is trained to roll a marble—initially placed at a random position—toward the center of the GelSlim sensor image. The policy is trained using supervised learning on GelSlim tactile data and learns to guide the marble using the tactile feedback provided by the sensor.

% We then test the transferability of this policy to a different sensor, DIGIT, by using our T2D2 model to generate the corresponding GelSlim tactile signal from the DIGIT input. This allows the original policy, trained exclusively on GelSlim data, to be applied to DIGIT inputs \textbf{without any retraining or modification}.

% The results show effective transfer: the policy succeeds in \textbf{15 out of 20 trials} when using DIGIT inputs translated to GelSlim, compared to \textbf{20 out of 20} using real GelSlim signals. This experiment demonstrates that our cross-sensor generation framework supports not only perception tasks but also simple sensor-conditioned control, enabling policy reuse across different tactile hardware.

\section{Appendix}
\subsection{Implementation Details}
\label{implementation_details}
\mypar{Diffusion Model}
Our implementation of the diffusion model closely follows Stable Diffusion~\cite{rombach2022high}, with the difference that we use a ResNet-50 to generate the GelSlim encoding from GelSlim images for conditioning. 

The model is optimized for 30 epochs by Adam~\cite{kingma2014adam} optimizer with a base learning rate of $10^{-5}$. The learning rate is scaled by $\text{gpu number} \times \text{batch size}$. We train the model with batch size of 48 on 4 NVIDIA A40 GPUs. 

At inference time, the model conducts 200 steps of the denoising process with a $2.54$ guidance scale.

\mypar{VQ-VAE}
We use a VQ-VAE architecture similar to the one proposed by Van den Oord et al \cite{van2017neural} for the style transfer. Before training VQ-VAE, we processed the sensor images by obtaining the difference between the deformed image at the moment of contact with the object and the undeformed image. In addition, we resize these images from the sensor images' original size to 128x128 and keep their corresponding numbers of channels.  The input to our model is a 128x128 subtracted Gelslim RGB image, and the output is the corresponding 128x128 subtracted depth map Soft Bubbles image. The input image $x$ is passed through a CNN encoder to generate a vector in the latent space $z$. This latent vector is then quantized via a collection of discrete vectors known as the \emph{codebook}, such that $z_e(x)$ is transformed into $z_q(x)$. This quantized latent vector is passed through a CNN decoder to generate the final image $\Tilde{x}$. The encoder parameters, quantization codebook vectors, and decoder parameters are all learned such that mean squared-error in the latent space quantizations and output reconstructions are minimized.

\subsection{Depth Esimation Details}
We map a tactile image \( I_S \) to a depth map \( D_S' \) and a contact mask \( M_S' \). For the depth map, we minimize the scale-invariant logarithmic loss~\cite{eigen2014depth,depth_anything_v2}:
{\small
\begin{equation}
\mathcal{L}_{\text{silog}} = \sqrt{ \frac{1}{n}\sum_{i=1}^{n} \left( \log D_{S,i} - \log D_{S,i}' \right)^2 - \lambda \left( \frac{1}{n}\sum_{i=1}^{n} \left( \log D_{S,i} - \log D_{S,i}' \right) \right)^2 },
\label{eq:silog_loss}
\end{equation}}%
which measures the discrepancy between the logarithms of the ground truth depth \(D_S\) and the estimated depth \(D_S'\). We simultaneously train the model to estimate the contact mask, $ M_S'$, using binary cross entropy loss.

\subsection{Depth Adaptation Stage Details}
For the adaptation stage, we use the depth map \(D_S'\) and its corresponding binary mask \(M_S'\) from the source sensor tactile image \(I_S\) obtained with our Depth Estimation Model to find the equivalent depth map \(D_T''\) and mask \(M_T''\) as if the target sensor was in contact.

% Mention that we also need the K_T instrinsic matrix and M_T (mesh) geometry of target sensor and T_S->T (Add alignment frame explanation here and) 

First, we define the set of valid pixel coordinates, using the contact mask \(M_S'\), as follows:
\begin{equation}
\Omega = \{ (u,v) \mid M_S'(u,v) = 1 \}
\end{equation}

For each valid pixel \((u,v) \in \Omega\), we back-project the depth value into 3D space using the inverse intrinsic matrix \(K_S^{-1}\) and subsequently transform the resulting 3D point from the source sensor frame to the target sensor frame using the rigid transformation \(T_{S \rightarrow T}\). Here, \(T_{S \rightarrow T}\) is defined as the composition of the transformation from the source sensor to the alignment frame, \(T_{S \rightarrow A}\), and from the alignment frame to the target sensor, \(T_{A \rightarrow T}\); that is, \(T_{S \rightarrow T} = T_{A \rightarrow T} \circ T_{S \rightarrow A}\).The alignment frame is defined as a common reference frame shared across different tactile sensors, representing the same touch event. This standardization allows for consistent interpretation of the contact geometry, regardless of each sensor's unique geometry or pose. This process is expressed as follows.

\begin{equation}
    \mathcal{P}_T = \left\{ T_{S \rightarrow T} \left( D_S'(u,v) \cdot K_S^{-1} \begin{pmatrix} u \\ v \\ 1 \end{pmatrix} \right) : (u,v) \in \Omega \right\}.
    \label{eq:estimated_target_pcd}
\end{equation}

Where, \(\mathcal{P}_T\) represents the resulting point cloud in the target sensor frame. Then we find the target sensor depth map and its mask, defined as:

\begin{equation}
    D_T''(u,v) = 
    \begin{cases}
      Z, & \text{if there exists } p = (X,Y,Z)^\top \in \mathcal{P}_T \text{ such that} \begin{pmatrix} u \\ v \\ 1 \end{pmatrix} = \frac{1}{Z}\,K_T\,p \\[1mm]
         &  \text{ and } (u,v) \text{ is within the image bounds}, \\[1mm]
      0, & \text{otherwise}.
    \end{cases}
    \label{eq:target_depth_projection}
\end{equation}

\begin{equation}
M_T''(u,v) =
\begin{cases}
1, & \text{if there exists } p \in \mathcal{P}_T \text{ such that } p \text{ projects to } (u,v) \\
   & \quad \text{ and } \mathrm{SDF}(p, \mathcal{M}_{\text{target}}) \leq 0, \\
0, & \text{otherwise.}
\end{cases}
\label{eq:mask_target_uv}
\end{equation}

\begin{figure}[b]
    \centering
    \includegraphics[width=0.85\linewidth]{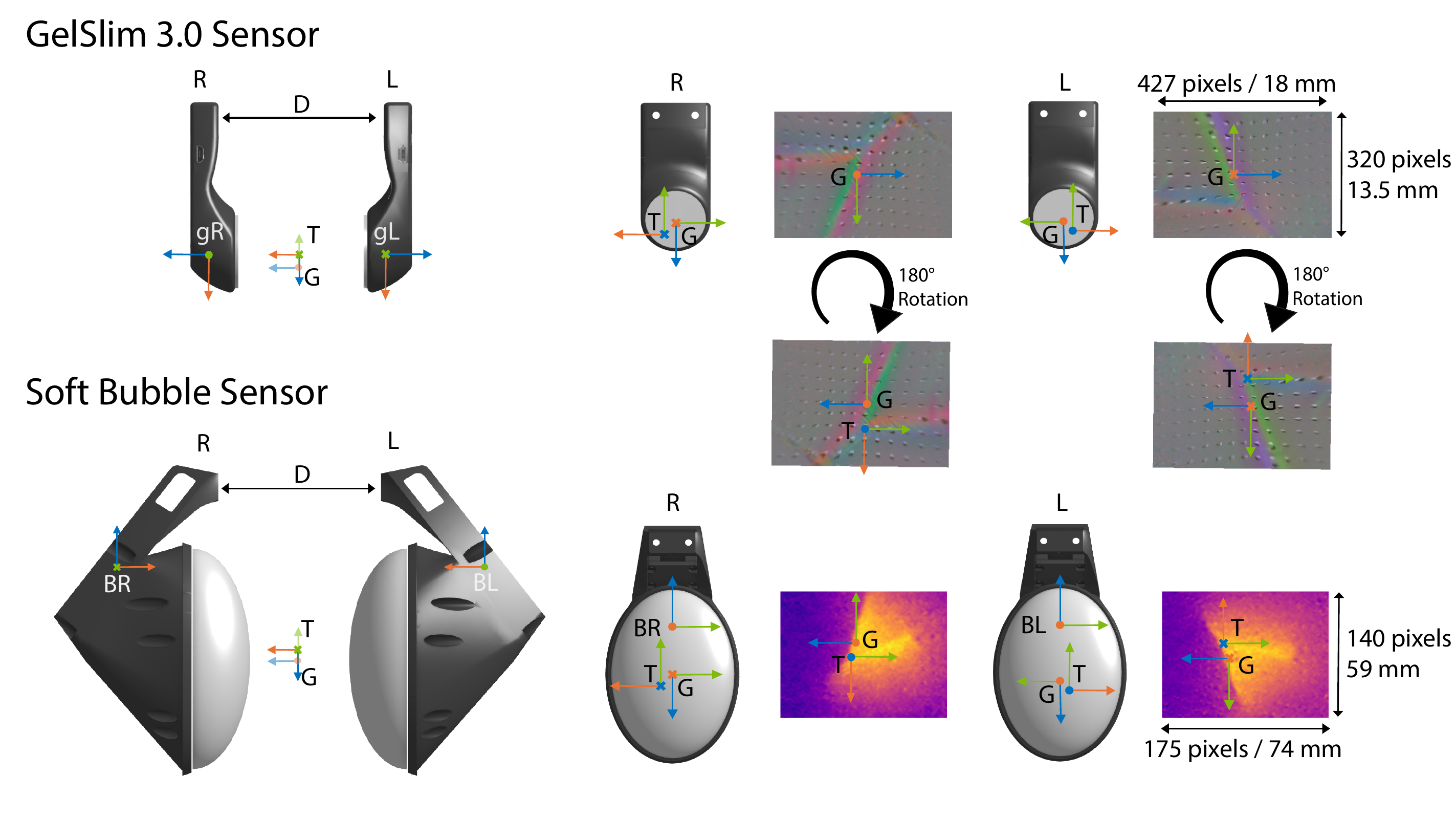}
    \caption{\textbf{GelSlim and Soft Bubble Alignment.} This figure shows each sensor's main coordinate frames: GelSlim camera frames (gR, gL), Soft Bubbles camera frames (BR, BL), grasp frames (G), and tool frames (T). D corresponds to the distance we keep between the same type of sensors during data collection. We can see the grasp frames projection in the image plane of each sensor and notice the need to rotate for alignment. For each coordinate frame, the x-axis is shown in red, the y-axis is shown in green and the z-axis is shown in blue.}
    \label{fig:sensor_details}
\end{figure}

\subsection{Sensor Alignment Details}
\label{sensor_details}
Figure \ref{fig:sensor_details} shows each sensor's main coordinate frames: GelSlim camera frames (gR, gL), Soft Bubbles camera frames (BR, BL), and grasp frame (G). For our Touch2Touch dataset, we align both sensors by locating the grasp frame of each at the same section of the manipulated object to obtain paired tactile signatures. In addition, we keep both sensors at a distance D, shown in Figure \ref{fig:sensor_details}. The final step to align the tactile signatures is to rotate one sensor image by 180$^{\circ}$. This rotation is necessary for the grasp frames to be aligned. We can see the grasp frames projection in the image plane of each sensor in Figure \ref{fig:sensor_details} and notice the need to rotate for alignment. This figure also shows the difference in size between the sensors' images in pixel space and millimeters.

\subsection{Data Collection Tools}
\label{data_collection_tools}
\begin{figure}[h]
    \centering
    \includegraphics[width=1.0\linewidth]{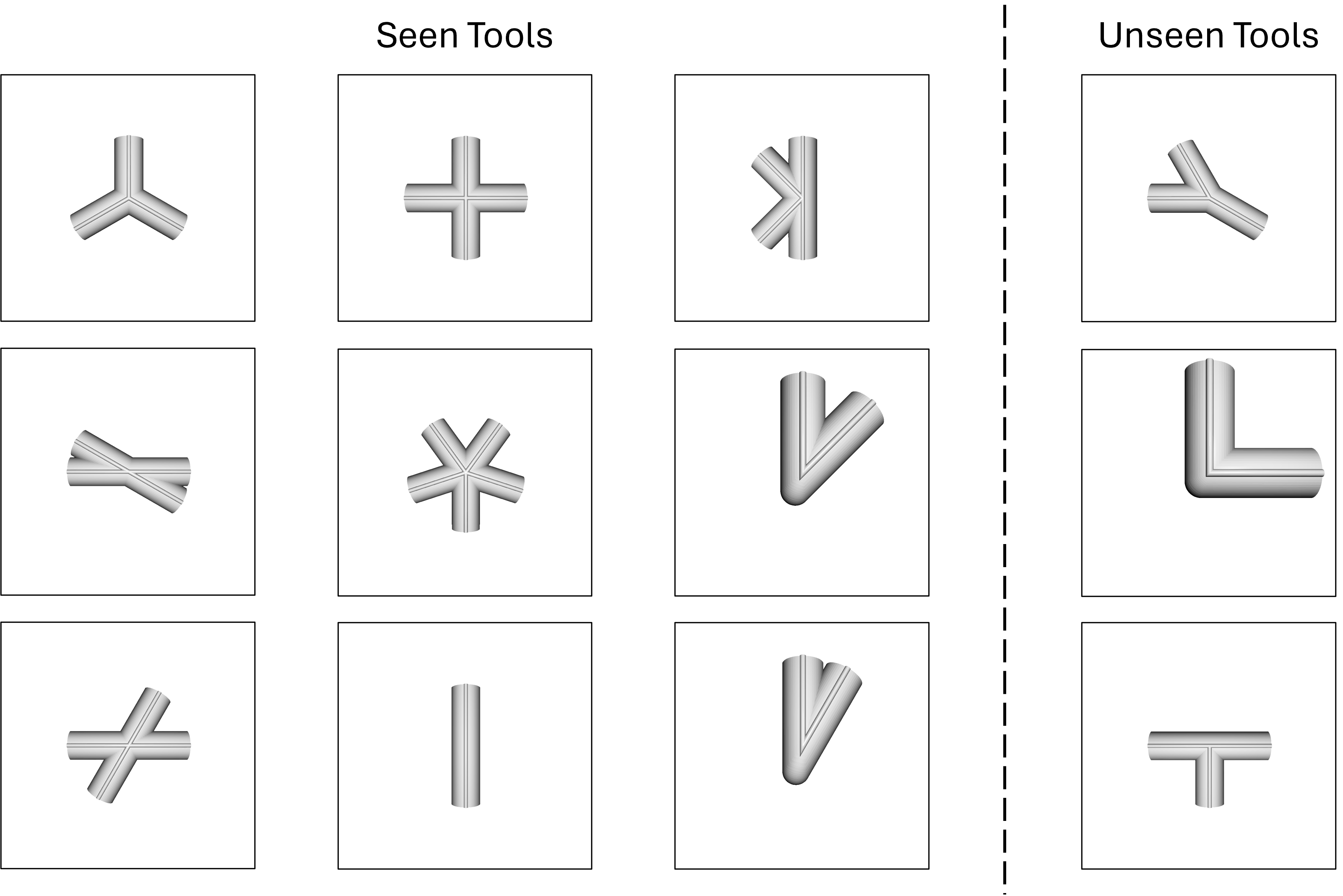}
    \caption{\textbf{Dataset Tools.} The left side of the image shows the geometries of the tools that were seen during training of the generative models. The right side of the image shows the geometries of tools that were not seen during training of the generative model. }
    \label{fig:dataset_tools}
\end{figure}

Figure \ref{fig:dataset_tools} shows the geometries of both the seen and unseen tools during the training of generative models. The unseen tools are designed to contain geometric features that are distinct from those of the training tools.

\end{document}